\documentclass{jimis-final-en}
\usepackage{array}
\usepackage{pgfplots}
\usepackage{graphicx}
\usepackage[utf8]{inputenc}

\title{Contextualizing Geometric Data Analysis and Related Data Analytics: \\
A Virtual Microscope for Big Data Analytics}
\author[*1]{Fionn Murtagh}
\author[2]{Mohsen Farid}
\affil[1]{University of Huddersfield, UK}
\affil[2]{University of Derby,UK}
\corrauthor{fmurtagh@acm.org}

\doi{10.18713/JIMIS-010917-3-1}{}
\review{12/2/2016}{6/2/2017}
\publication{3}{2017}

\issue{Digital Contextualization}
\editors{Frédéric Lebaron, Brigitte Le Roux, Fionn Murtagh, Evelyn Ruppert}

\begin{document}

\maketitle



\abstract{The relevance and importance of contextualizing data analytics is described.
Qualitative characteristics might form the context of quantitative analysis.  Topics
that are at issue include: contrast, baselining, secondary data sources, 
supplementary data sources; dynamic and heterogeneous data.  In geometric data 
analysis, especially with the Correspondence Analysis platform, various case studies are 
both experimented with, and are reviewed.  In such aspects as paradigms followed, 
and technical implementation, implicitly and explicitly, an important point made is
the major relevance of such work for both burgeoning analytical needs and for new 
analytical areas including Big Data analytics, and so on.  For the general reader, 
it is aimed to display and describe, first of all, the analytical outcomes that are
subject to analysis here, and then proceed to detail the more quantitative outcomes 
that fully support the analytics carried out.} 


\section{Introduction}

\strut
\vspace{-4ex}

Clearly, regression and other supervised, predictor and predicted relationships,
have their fixed roles in all of data analytics.  In unsupervised analytics, 
information discovery is at issue, from what can be multi-faceted data sourcing.
It is this latter, unsupervised analytics, which is our focus, our interest, 
and our motivation.  We begin with a most general introduction to how context
is sourced from, or accompanying, our measured data.  Our measured, and therefore
quantitative data may be in some framework or other which is assessed qualitatively.
Perhaps our measured observations are to be contrasted or baselined against 
what might be termed secondary data sources, or supplementary data sources.  

It may be a case of dynamic (evolving) data and also data of heterogeneous precision 
(sensitivity, trustworthiness). Also, at issue may be both qualitative and 
quantitative data.  

Examples of such multi-faceted data sources are the increasing
prevalence of secondary data sources in the domains of Big Data, and Internet of Things. 
Such secondary data sources may well count as contextual data.  

Given multi-faceted
data, it is, or it can be, open to the data scientist to consider different 
contextualization strategies.  Consider e.g.\ how mobile communications, and their 
monitoring, provide locational data on all device bearers.  So, for example, 
transport analytics can be contextually based on such secondary data and information
sources.  

In this article, it is sought to explore general strategies for 
contextualizing analytics.  Technically, the main attributes are aided, for 
analytical interpretation, i.e.\ extracting information from data, and consolidating
that information into knowledge, by the supplementary, contextual, attributes.  

The general discussion of \cite{wessel} 
points to the need for focus in benefits 
to be drawn from data analytics.  In a general sense, we may regard dimensionality 
reduction as a form of focused data analytics, since typically (in Correspondence
Analysis, or Principal Components Analysis, PCA, or multidimensional scaling, etc.) 
the percentage variance 
or inertia explained by the reduced dimensionality space becomes the focus.   
Similarly feature selection is often used, to focus on the selected variables.  
Often the clustering of data is with the objective of retaining the clusters for
the analytics.  Therefore summarization is associated with focusing the analysis.  
Summarization is a most major theme in data analytics, through information space
dimensionality reduction, through visualization, and through basic and core statistics.

Thus focus is a key practical consideration in data analysis.  Then, though, 
Jean-Paul Benz\'ecri's principle of what we are analysing being both homogeneous 
and exhaustive should be very central also, from a practical perspective 
(\cite{brigitte2014}, p.\ 9).  Just in this practical 
sense, we begin by regarding contextualization as the association of two or more 
foci, i.e.\ two or more directions of analytical interest.  This association is
asymmetric.  One focal point is located with reference to the other.  This can be 
generalized to multiple focal directions in our analyses.  

We begin with such contextualization in general.  Then we take contextualization 
further in the context of Big Data.  
In the former, applications related to health and lifestyle analytics are at issue.
In the latter, a central theme is the use of secondary data, whether or not 
with Big Data characteristics, so as to direct and focus the analytics.

\section{Short Introduction to General Contextualization}
\label{sect2}

Correspondence Analysis, CA, is most appropriate for latent semantic analysis, 
with input data being a cross-tabulation of observations or individuals, 
and variables or attributes.  The cross-tabulation is usually frequency of 
occurrence data, encompassing presence-absence values.  There can be considered
also recoded quantitative data.  Quantitative data, without recoding, can 
be used.  Therefore CA is very appropriate for mixed qualitative and quantitative
data.  Categorical data is quite central here.  Contrasted with PCA, where attribute 
centering to zero mean, and attribute
reduction to unit standard deviation, termed standardization when both carried out, 
and contrasted with the TF-IDF, term frequency, inverse document frequency that is
used in Latent Semantic Analysis of document-term cross-tabulation data, CA 
uses the following.  The chi-squared distribution is defined on both rows and 
on columns.  This is a weighted Euclidean distance of, respectively, row and 
column profiles.  A profile is defined as the row or column vector being divided,
respectively, by the row or column total.  This is what results from frequencies
in the input cross-tabulation matrix, i.e.\ each value divided by the grand total. 

In \cite{murtagh2005,murtagh2017}, 
CA is characterized as ``a tale of three 
metrics'', these
being the chi-squared metric for the dual clouds of observations and attributes, 
the factor space endowed with the Euclidean distance, and from that, 
hierarchical clustering that is a mapping into an ultrametric topology.  

Multiple Correspondence Analysis, MCA, typically works on the set of modalities
of the variables. When resulting from a questionaire's questions, and with just 
one response for each respondent from among the question's modalities, then an 
interesting result of that is that the row (i.e.\ respondent or individual) 
masses, defined from the row totals, are constant.  It then results that the 
chi-squared metric between row profiles becomes a weighted Euclidean distance.

\subsection{Baselining or Contextualizing Analysis}

In \cite{solene}, 
there is an important methodological 
development, concerning statistical inference in Geometric Data Analysis,
i.e.\ based on MCA.  At issue is statistical
``typicality of a subcloud with respect to the overall cloud of individuals''.
Following an excellent review of permutation tests, the data is introduced:
6 numerical variables relating to gait, body movement, related to the following;
a reference group of 45 healthy subjects; and a group of 15 Parkinsons illness
patients, each before and after drug treatment.  \cite{brigitte2014} 
(section 11.1) 
relates to this analysis, of the, in total, 45 + 15 + 15 observation vectors, 
of subjects between the ages of 60 and 92, of average age 74. 

First there is 
correlation analysis carried out, so that when PCA of standardized variables 
is carried out, it is the case that the first two 
axes explain 97\% of the variance.  Axis 1 is characterized as ``performance'', 
and axis 2 is characterized as ``style''.  Then the two sets of, before 
treatment, and after treatment, 15 Parkinsons patients are input into the 
analysis as supplementary individuals.  \cite{solene} 
is directly
addressing statistically the question of effect of treatment.  Just as in 
\cite{brigitte2014}, 
the healthy subjects are the main individuals, and the 
treated patients, before and after treatment, are the supplementary individuals. 
This allows to discuss the subclouds of the before, and of the after treatment 
individuals, relative to the first, performance, axis, and the second, style, axis. 
The test statistic, that assesses statistically the effect of medical treatment 
here, is a permutation-based distributional evaluation of the following 
statistic.  The subcloud's deviations relative to samples of the reference cloud
are at issue.  The Mahalanobis distance based on covariance structure of the 
reference cloud is used.  The test statistic is the Mahalanobis norm of 
deviations between subcloud points and the mean point of the reference cloud. 

In summary, this exemplifies in a most important way, how supplementary 
elements and the principal elements are selected and used in practice.  The 
medical treatment context is so very clear in regard to such baselining, i.e.\
contextualizing, against healthy reference subjects.

\section{Data: Mental Health: Adult Psychiatric Morbidity Survey, England, 2007}
\label{sect5}

In \cite{coenders}, 
there is this motivation for MCA, 
in the Abstract: ``Most of the relevant factual and subjective items in quality of 
life questionnaires are qualitative and call for a multiple correspondence analysis
type of analysis.''  This is an important vantage point on the analytics: 
``Subjective well-being questions play the role of active variables and objective
well-being questions that of illustrative variables.''  It is noted how MCA is able
to handle non-responses and also interviewer effects.  Concluded is: ``The analysis
was carried out without accounting for non-response and interview effects and the 
interpretation of the axes became much less clear.''  The first two axes of factors
are found to account for a polarisation between high quality of life, and low
quality of life.  The data at issue in this work, \cite{coenders}, 
is a survey of
elderly people, over 65 and over 75, carried out in Girona, Spain, in 1999.  There 
were 2000 subjects.  The questions comprised background variables (location, age, 
gender, last job, family), factual health questions, and a subjective (perception) 
health question.  Also included were questions on activity, movement, mobility; and  
housework, shopping, managing one's income.  Modalities were specified for each 
question.  Social satisfaction questions had Likert scale modalities.  

Is is noted in \cite{coenders} 
how missing data may be ``substantively 
interpretable'', and will require a large number of presences in the missing value 
modalities.  The overall presence of a sufficiently large number of missing 
values is so as not to overly impact the analysis through factors being found
for the rarely occurring missing values.  It is also noted how impactful, in the 
analysis, missing values will be if they are common to clusters of variables.  
Ways to address this include doubling the data variables, and perhaps even modalities
of the variables, see \cite{murtagh2005,coenders}, 
doubling the column set, 
so that it results that row sums remain constant.  It is even possible under 
such circumstances to modify the ``deflated'' and sufficiently large eigenvalues
(\cite{jpb1979,greenacre1993,coenders}).

Quite a unique analytics approach is pursued by \cite{coenders}, 
firstly availing 
of MCA, with appropriate data coding, being equivalent to PCA, 
with appropriate data normalization.   Secondly, correlation analysis, 
and analysis of variance, are carried out on the principal factor plane, accounting 
for most of the variance, and also k-means clustering.  In summary, on the given 
data, the work of \cite{coenders} 
is well accomplished.  

In the following, now, we will take as a case study, survey data that is 
heterogeneous and diverse.  We note that for the present initial study of 
this data, we will place ``Don't know'' responses, with very low response totals,
as supplementary response modalities, that, thus, will not influence detrimentally 
the essential aspects of our analytics.  

A periodic survey of mental health is used.  This is a continuing survey of 
adult psychiatric morbidity in England, carried out as a household survey, 
\cite{HSCIC}.  
The number of respondents is 7403.  There are 1704 
variables, including questioning of the subjects about symptoms and disorders,
psychoses and depression characteristics, anti-social behaviours, eating 
characteristics and alcohol consumption, drug use, and socio-demographics, 
including gender, age, educational level, marital status, employment status, 
and region lived in.

We begin with a general, visualization-based, presentation of the outcome of this
quite directed analysis.  Following the description of the principal factor plane 
display, there will be a brief presentation that justifies the visual, i.e.\ 
display, findings.  

As a first analysis, the following variables were selected: 14 questions, hence
14 categorical variables, relating to ``Neurotic symptoms and common mental 
disorders''.  These are described in \cite{HSCIC} 
(Appendix C). 
Almost all of these
variables had as question responses, whether or not there were symptoms or disorders
in the past week, one question related to one's lifetime, and one question related
to the age of 16 onwards.  Another question set was selected, relating
to socio-demographic variables, noted in the foregoing paragraph.  In this set of 
socio-demographic variables, there were 9 variables.  

\begin{figure*}[h]
\centering
\includegraphics[width=10cm]{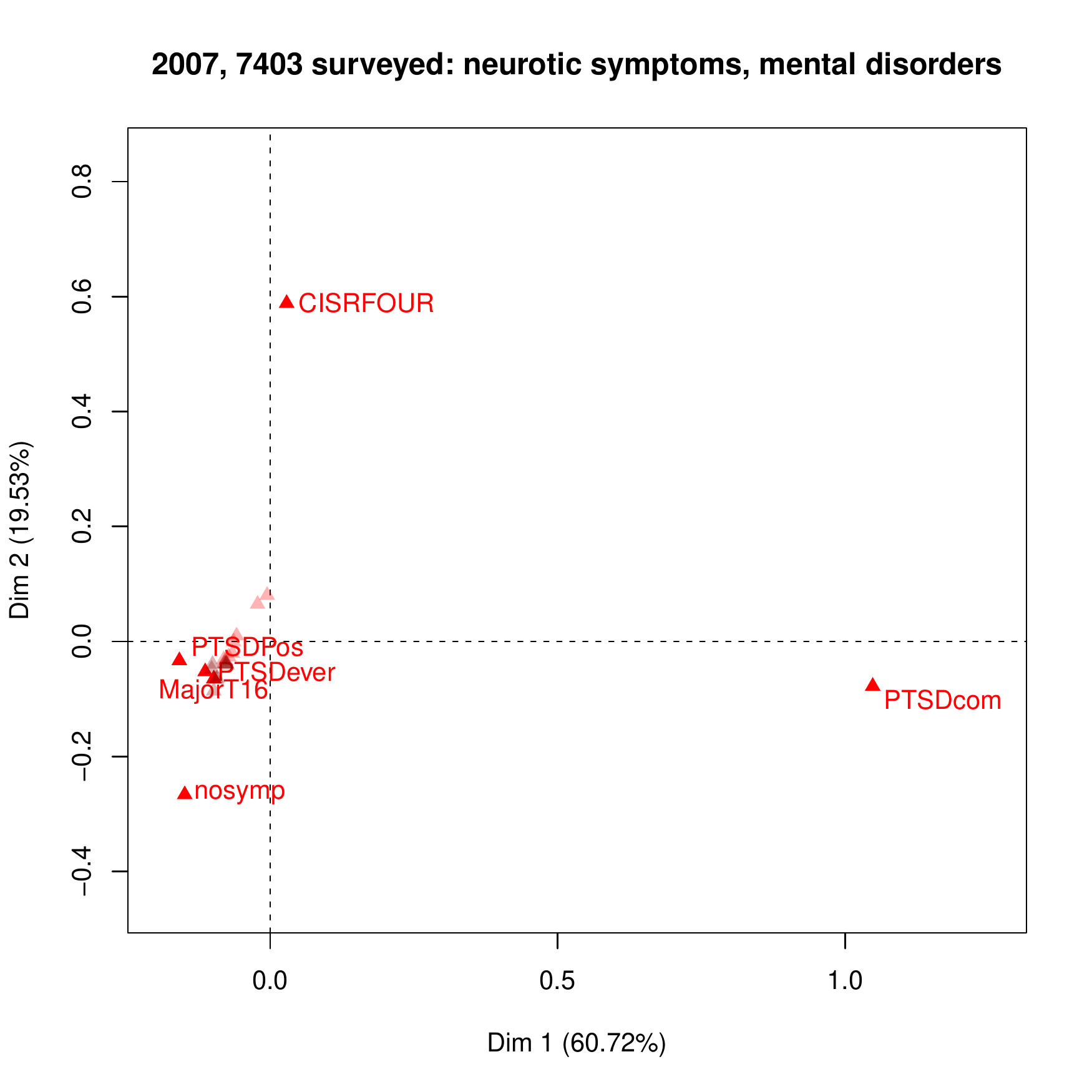}
\caption{Adult psychiatric morbidity survey 2007, England, household survey. 
The socio-demographic variables, as supplementary variables are close to the origin.
Displayed are the 6 highest contributing variables.}
\label{fig1}
\end{figure*}

An initial display of the neurotic symptoms and common mental disorders, see
Figure \ref{fig1}, sought to have socio-demographic variables as supplementary.  But
these were projected close to the origin, therefore showing very little 
differentiation or explanatory relevance for the symptoms and disorders data.  

In Figure \ref{fig1}, it is found that factor 1 is explained as PTSDcom, ``Trauma 
screening questionnaire total score'' versus all other variables.  Factor 2 is 
explained as ``CISRFOUR'' versus ``nosymp''.  These are, respectively, 
``CIS-R score in four groups, 0-5, 6-11, 12-17, 18 and over. (CIS-R = Common Mental 
Disorders questionnaire)''; and no neurotic symptoms in the past week.  

What has been stated about Figure \ref{fig1}, from the visual interpretation is 
fully supported by the following.  The contribution of ``CISRFOU'' to factor 2 
is 79.09\%, and its correlation, viz.\ cosine squared, with factor 2 is 0.96. 
The contribution of ``PTSDcom'' to factor 1 is 89.65\%, and its cosine squared 
with factor 1 is 0.99.  The cosines squared of ``PTSDever'' with factor 1 is 0.26;
that of ``MajorT16'' with factor 1 is 0.21; and that of ``PTSDPos'' with factor 1 
is 0.71.  These last three questions are close to the origin, i.e.\ very average 
and unexceptional.  Finally, in this figure, ``nosymp'' has this contribution to 
factor 2: 13.39\%, and its cosine squared with factor 2 is 0.40.  

To summarize the analysis at issue here, the following are the data characteristics:
7403 individual respondents; 23 questions with 107 response modalities; of these 
questions, 1 to 14 are active, and 15 to 23 are supplementary.  The 14 active 
questions have 52 response modalities, and the 9 supplementary questions have 55
response modalities.  As noted above, the 14 active questions were relating to 
``Neurotic symptoms and common mental disorders'', and the 9 supplementary questions
were socio-demographic variables.  

\begin{figure*}[h]
\centering
\includegraphics[width=10cm]{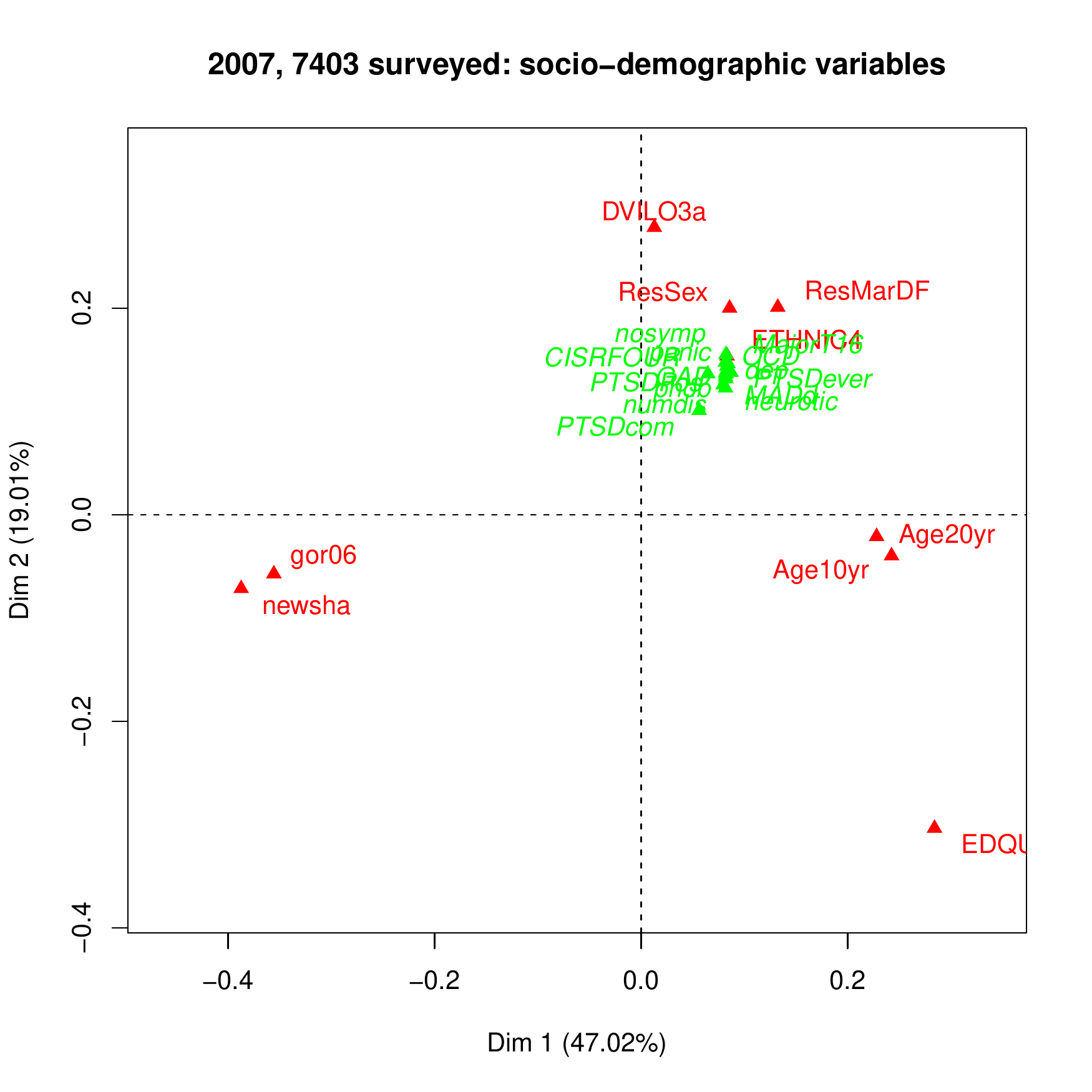}
\caption{Adult psychiatric morbidity survey 2007, England, household survey.
The neurotic symptoms and common mental disorders, as supplementary variables are
close to the origin.} 
\label{fig2}
\end{figure*}

Next, Figure \ref{fig2}, it was sought to characterize the socio-demographic data, 
and then to see if the neurotic symptoms and common mental disorders data could be 
explanatory and contextual for the socio-demographic data.  But no differentiation 
was found for these supplementary variables, indicating no particular explanatory 
capability in this particular instance.  

It may be just noted how, in Figures \ref{fig1} and \ref{fig2}, the main 
and supplementary variables were interchanged.  Respectively, the symptoms 
and demographic variables were main and supplementary; then the main and supplementary 
variables were the demographic variables and symptoms.  This was done in order
to explore the data.  

In Figure \ref{fig2}, factor 1 is seen to have age and education level counterposed 
to home region.  Factor 2 is seen to have educational level counterposed to: employment 
status, gender, marital status, ethnicity.  (In the paragraph to follow, 
the names of some of these variables are explained.) 

Just to complete and confirm, numerically, the contributions to the axes, then validated
by the cosine squared, this is as follows for these questions.  We note once again, 
that this analysis is carried out on the 9 socio-demographic questions at issue here, 
and as 14 supplementary questions, these were questions related to neurotic symptoms 
and common mental disorders. On the negative half-axis of factor 1, ``newsha'', ``gor06'' 
(see the next paragraph for the names of these questions) had contributions, respectively,
of 35.5\% and 27.8\%.  Their cosines squared were, respectively, 0.96 and 0.97.  The main
contributions for the positive half-axis of factor 1 were: ``EDQUAL5'' and ``Age10Yr'', 
with contributions of 16.2\% and 10.4\%. Their cosines squared were 0.38 and 0.40.  The next
contribution to factor 1, less than all these values, was 5.24\% and its cosine squared was
0.42.  This was the question ``Age20yr''.  (The questions ``Age10Yr'' and ``Age20Yr'' were 
both querying the individual's age, with 10-year and with 20-year intervals for age.) 

For factor 2, in  Figure \ref{fig2}, the highest contributions are not at all as high as 
for factor 1, and nor are the cosines squared.  The contribution to factor 2, of 
``EDQUAL5'', negatively positioned for its coordinate on factor 2, is 45.8\%.  The cosine
squared of ``EDQUAL5'' with factor 2 is 0.43.  Smaller than these values of contributions, 
and cosines squared is the collection of active variables, displayed in  Figure \ref{fig2}, 
with coordinates on the positive half axis of factor 2. 

\begin{figure*}[h]
\centering
\includegraphics[width=10cm]{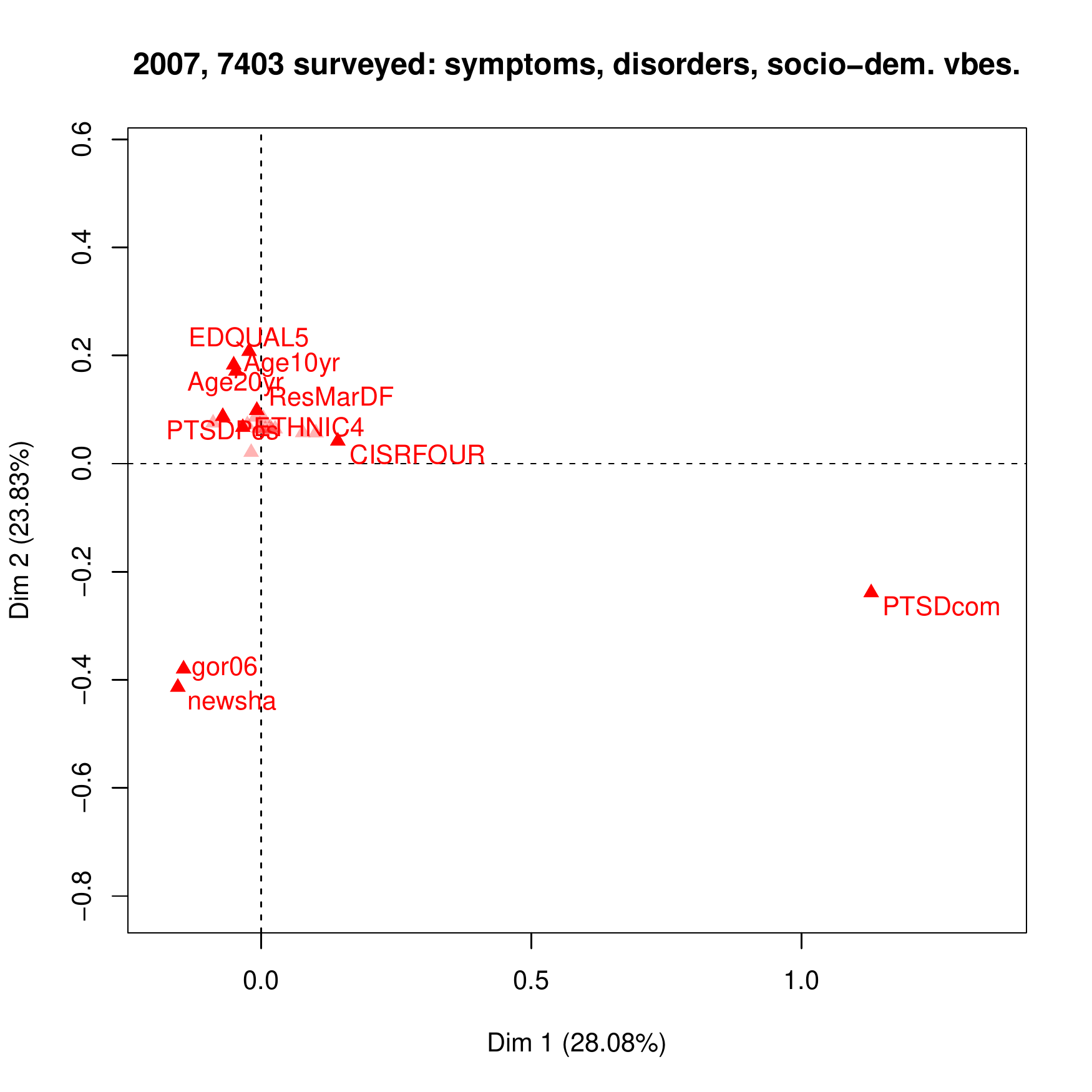}
\caption{Adult psychiatric morbidity survey 2007, England, household survey.
The analysis has the neurotic symptoms and common mental disorders, and the 
socio-demographic variables.  Displayed are the 10 highest contributing variables 
to the principal plane.}
\label{fig3}
\end{figure*}

Finally, it was checked whether neurotic symptoms and common mental 
disorders data should be jointly analysed with the socio-demographic data.  Figure 
\ref{fig3} shows this outcome.  On the positive first factor, what is 
particularly important, from the contribution to the inertia, is PTSDcom, 
``TSQ (Trauma Screening Questionnaire) total score''.  The negative second factor, 
is highly influenced by these two variables: ``gor06'', ``newsha'', respectively: 
``Government office region'' and ``Strategic Health Authorities''.   These were both
sets of geographic regions in England, respectively 9 and 10.  (The 7th such region 
in each of these variables was London.)  

One aim arising out of this analysis is to see if we can formulate forms and 
expressions of ``mental capital''.  This concept is noted in section \ref{sect21}
where an example of it was going for a walk, therefore taking exercise,
in the countryside or parkland.  


A summary, and preliminary, interpretation derived from Figure \ref{fig3} is how
factor 1 accounts for recorded trauma, and factor 2 accounts for region of the
respondent.  

\section{Analysis of Activities, Contextualized for General Health, Depression and 
Demographics}
\label{sect4}

Our primary aim in this initial analysis of one year's data is to seek insightful 
properties in the data, and also to become aware of where there is a lack of 
impressionable outcomes.  

We base our analysis of (i) general health, (ii) activity, (iii) depression, and 
(iv) demographics, on the following: activity.  We do so for two reasons, 
firstly that in the future sensor systems, related to the surveillance or to the 
Internet of Things, such systems will give rise to a very great amount of free and 
flexible data, that is secondary and contextual in its essential nature.  The 
second reason is to pursue observation, associated with a line of enquiry, towards 
determining forms and expressions of mental capital.  As background and definition 
of mental capital, 
the following may be considered, from \cite{cooper}:  
``... paths to wellbeing -- 
being active, taking notice, connecting with others, giving and learning -- can be 
achieved in the outdoors and can lead to further resilience, enhance self-esteem, 
improve learning and develop emotional intelligence, thus building mental capital.''
This short account, \cite{cooper}, 
entitled ``The `mental capital' value of the 
outdoors'', begins as follows: ``It is now well known that mental wellbeing is 
linked with people using outdoor environments and feeling connected to nature.'' 
Here we seek to pursue what can be forms of mental capital at the elementary stages
of motion and movement.  A further short essay which is after that of 
\cite{cooper} 
is entitled ``Walking improves mental wellbeing''.

We have 7403 respondents, or individuals. 
Now, from the 57 questions, furnishing variables, these are as follows: 
general health: 19, activities: 7, depression: 24, and demographics, 7.  These 
questions are, in the set of 57, and in succession, as follows: 1 to 19, 
20 to 26, 27 to 50, and 51 to 57.  The numbers of modalities for these question 
sets, viz.\ for respectively general health, activities, depression and demographics,
are: 101, 25, 154 and 44.  In total, the number of modalities is 324.
 
For activities, as noted there are 7 questions, i.e.\ variables.  The respective 
number of modalities are: 3, 4, 3, 4, 4, 3, 4.  In total, there are 25 modalities.
Let us look in turn at these modalities: ``No, no difficulty at all'', 
``Yes, a lot of difficulty'', ``Yes, some difficulty''; 
``Don't know'', ``No, no difficulty at all'', ``Yes, a lot of difficulty'', 
``Yes, some difficulty''; ``No, no difficulty at all'', ``Yes, a lot of difficulty'',
``Yes, some difficulty''; ``Don't know'', ``No, no difficulty at all'', 
``Yes, a lot of difficulty'', ``Yes, some difficulty''; ``Don't know'', 
``No, no difficulty at all'', ``Yes, a lot of difficulty'', ``Yes, some difficulty''; 
``No, no difficulty at all'', ``Yes, a lot of difficulty'', ``Yes, some difficulty'';
and ``Don't know'', ``No, no difficulty at all'', ``Yes, a lot of difficulty'', 
``Yes, some difficulty''.  Due to very low total frequencies, we find that the four 
cases of ``Don't know'' have very few respondents.  Given hundreds or thousands 
of respondents responding to the modalities of the other questions, for these 
modalities, the 4th, 11th, 15th and 22nd, we find the number of respondents to them 
to be, respectively, 1, 2, 16 and 2.  Very small numbers of respondents, like we have
here, leads to exceptionality and to over-influence on the Correspondence Analysis 
determining of new axes.  Hence we can make them (the four modalities, the 4th, 11th, 
15th and 22nd) as supplementary in the analytics that crosses respondents by 
retained modalities.  

In brief, these activity questions are related to the following summary forms: 

AcDif1: personal care such as dressing, washing oneself.

AcDif2: using transport.

AcDif3: medical care.

AcDif4: housekeeping, shopping.

AcDif5: gardening, household repair.

AcDif6: paperwork.

AcDif7: managing money.

\noindent
The activities data, itself to begin with, crosses 7403 respondents and 21 
response modalities for the 7 questions.  Of the 25 response modalities, 21 are
main attributes and 4, as explained above, are supplementary.  The eigenvalues 
as rates of inertia explain 26.8\% of the first factor's, or new axis's, inertia, 
and 14.1 \% of the second factor's, new axis's, inertia.  Modifying the 
rates of inertia is well based on such reasons as the following: emphasizing 
relative importance of the axes; have an index of deviation from sphericity of 
the factor space cloud; and adjusting for the important role in the defining of 
factors that is played by the faint and minor information present in sparse input 
data, as is the case here.  Cf.\ \cite{brigitte2014}, 
page 252.  
Having $Q$ questions, here $Q = 7$, the average 
eigenvalue is $1/Q$. For the set of eigenvalues greater than the average, 
$\lambda_{\ell}$, these are modified as follows: $ ( Q/(Q-1) )^2 ( \lambda_{\ell}
- 1/Q )^2$.  For the activities data being mapped into the Correspondence Analysis 
factor space, we find the modified rate of inertia associated with the first and 
second factorial axes to be 88.7 \% and 11.1 \%.  Therefore in relative terms, 
the first two axes account for essentially all the information content in this 
activities data.  

\begin{table}[h]
\centering
\begin{tabular}{|lrr|} \hline 
Question     & Factor 1  &  Factor 1  \\ 
Modality     & Contrib.  &  Coord.    \\ \hline 
AcDif5YESPLUS & 13.4794768 & 2.0040969 \\
AcDif4YESPLUS & 12.9812467 & 3.6075176 \\
AcDif2YESPLUS & 12.4704065 & 3.0844644 \\
AcDif1YESPLUS & 9.0781425  & 4.0460224 \\
AcDif6YESPLUS & 8.6472558  & 2.6082064 \\
AcDif1YES     & 6.1438251  & 1.7048671 \\
AcDif7YESPLUS & 6.1242130  & 3.0997642 \\
AcDif3YESPLUS & 5.6879326  & 4.6516383 \\
AcDif4YES     & 4.7280714  & 1.3135295 \\
AcDif2YES     & 3.6031529  & 1.2322215 \\
AcDif3YES     & 3.2779799  & 2.2872765 \\
AcDif5NO      & 2.8541595  & -0.3836511 \\
AcDif4NO      & 2.2641684  & -0.3143501 \\
AcDif2NO      & 2.1136694  & -0.3033419 \\
AcDif6YES     & 1.7505882  & 0.8094317 \\
AcDif1NO      & 1.4251889  & -0.2437384 \\
AcDif6NO      & 1.3210564  & -0.2411598 \\
AcDif7YES     & 1.1431343  & 0.7646080 \\
AcDif7NO      & 0.4998090  & -0.1441357 \\
AcDif3NO      & 0.2736968  & -0.1030599 \\
AcDif5YES     & 0.1328260  & 0.1856692 \\ \hline 
\end{tabular}
\caption{Factor 1.  The positive half axis is dominated by the ``YESPLUS'', 
strongly positive responses, with coordinates $> 2$.  While not 
strongly influential, through having large contributions to the 
inertia, nonetheless it is seen that the negative 
half axis is associated with the ``NO'' responses.}
\label{table1}
\end{table}

From Table \ref{table1}, the first axis is characterized, in regard to the 
positive half axis,  by ``Yes, a lot of 
difficulty'', denoted by ``YESPLUS'' in the modality naming here.  This is
counterposed, in regard to the negative half axis, by the ``No, no difficulty 
at all'' modalities.  

From Table \ref{table2}, the second axis is characterized, 
in regard to the positive half axis, by  
``Yes'', here denoting ``Yes, some difficulty'' modalities.
The negative half axis is characterized by the ``YESPLUS'',
i.e.\ ``Yes, a lot of difficulty''.  The modalities labelled
``NO'' here, i.e.\ ``No, no difficulty at all'', are near 
zero in value on this second factorial axis. 

\begin{table}[h]
\centering
\begin{tabular}{|lrr|} \hline 
Question     & Factor 2  &  Factor 2  \\ 
Modality     & Contrib.  &  Coord.    \\ \hline 
AcDif4YES    & 1.779799e+01 & 1.848803011 \\
AcDif2YES    & 1.593868e+01 & 1.880100211 \\
AcDif1YES    & 1.094332e+01 & 1.650642759 \\
AcDif3YESPLUS & 9.780141e+00 & -4.424953058 \\
AcDif1YESPLUS & 8.985115e+00 & -2.920108539 \\
AcDif4YESPLUS & 6.401865e+00 & -1.837853577 \\ 
AcDif5YES     & 5.588667e+00 &  0.873694803 \\
AcDif6YES     & 5.495999e+00 &  1.040444434 \\
AcDif7YESPLUS & 4.508513e+00 & -1.929423524 \\
AcDif2YESPLUS & 3.312458e+00 & -1.153246126 \\
AcDif6YESPLUS & 2.998210e+00 & -1.114143773 \\
AcDif5NO      & 2.301451e+00 & -0.249922602 \\
AcDif7YES     & 1.657123e+00 &  0.667843905 \\
AcDif5YESPLUS & 1.207904e+00 &  0.435216938 \\
AcDif3YES     & 1.039831e+00 &  0.934553925 \\
AcDif4NO      & 8.698488e-01 & -0.141347606 \\
AcDif2NO      & 7.175949e-01 & -0.128221512 \\
AcDif1NO      & 2.754251e-01 & -0.077731493 \\
AcDif6NO      & 1.544416e-01 & -0.059818268 \\
AcDif3NO      & 2.502948e-02 &  0.022609359 \\
AcDif7NO      & 3.959342e-04 & -0.002942987 \\ \hline
\end{tabular}
\caption{Factor 2. The positive half axis is dominated by the ``YES'',
positive responses.  The negative half axis respresents the 
``YESPLUS''.  The ``NO'' responses are near zero on this axis.}
\label{table2}
\end{table}

It follows from Tables \ref{table1} and \ref{table2} that the 
lower right quadrant is associated with a lot of difficulty 
indicated in the responses, contrasted with the negative first
axis that is associated with no difficulty.  In the sense then
of the sourcing of mental capital, it is the negative, but near zero,
first factor that is most relevant and demonstrably important.

\begin{figure*}[h]
\centering
\includegraphics[width=10cm]{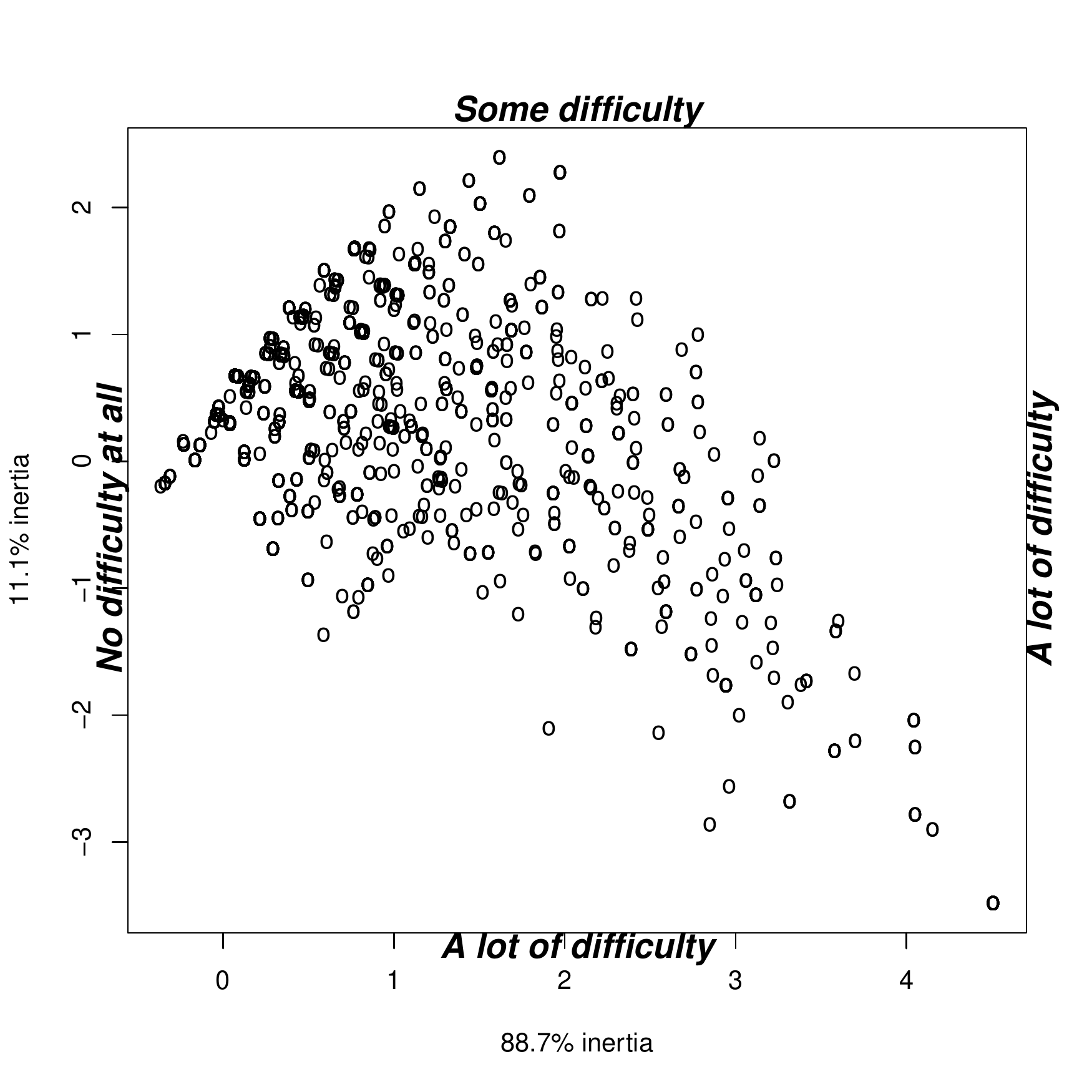}
\caption{Plotted are the 7403 respondents, shown with a ``o'' at their
locations.} 
\label{figcanew1}
\end{figure*}

Figure \ref{figcanew1} displays the principal factor plane with respondents
represented by an ``o'' point.  It would be easy to see which respondents 
had most difficulty in the activities at issue here, i.e.\ those who are
located towards the bottom of the bottom right quadrant.  

\begin{figure*}[h]
\centering
\includegraphics[width=10cm]{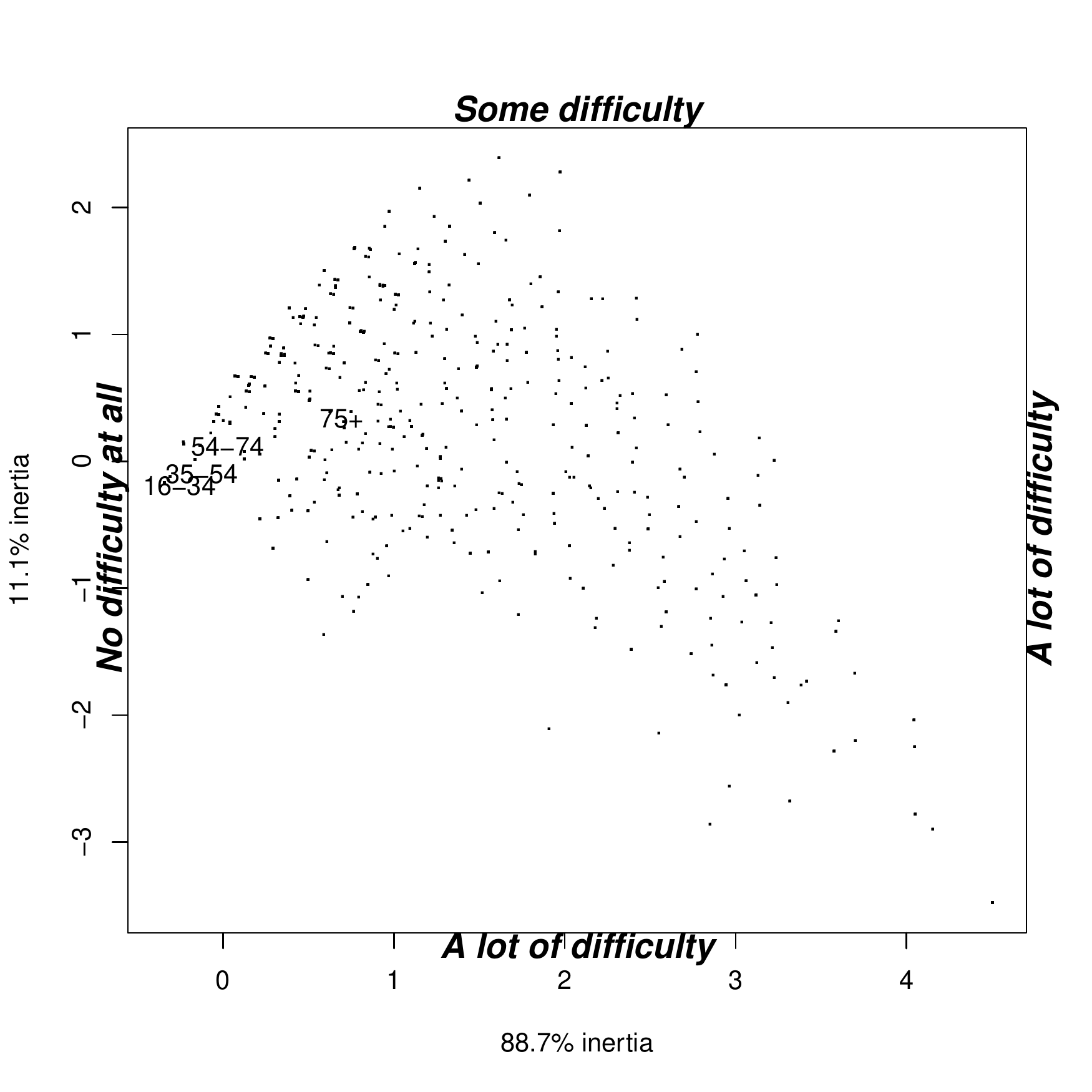}
\caption{As in Figure \ref{figcanew1}, locations of the respondents are shown 
here with dots.  Ages are supplementary elements.}
\label{figcanew2}
\end{figure*}

The modalities of all other questions can be taken as supplementary elements 
and mapped into the factor space.  Considering these supplementary modalities, 
``Economically inactive'', ``In employment'' and ``Unemployed'', these
were found to be to the left half axis, near zero, of factor 1.  Cf.\ Figure \ref{figcanew1}.
This same outcome was found for race attributes, ``South Asian'', ``White'', 
``Black''.  For age, Figure \ref{figcanew2} shows, to some extent, a worsening 
of activities with age $75+$.  Just to note are the numbers of respondents 
of the age intervals: 16--34, 1603; 35--54, 2543; 55--74, 2307; and 75+, 950. 
Given the theme of all of this work, psychiatric morbidity, we see that most ages
are quite average and neutral, in being close to the origin, with the exception,
to some extent of those aged 75+.  The coordinates of the 75+ mean projection, 
on axes 1 and 2 are 0.69 and 0.33, that are higher in value compared to the nearly
all negatively values coordinates of the other age intervals.  The cosine squared 
of this 75+ mean projection are, for axes 1 and 2, 0.07 and 0.02, and these very 
small values are nonetheless greater than all other ages.

For all the modalities of questions, their cosines with the very important and 
informative first factor were looked at, and the top ten cosines listed.  These modalities
of questions are as follows: 4 that were supplementary (being ``Don't know'' responses)
for the Activities question set; 19 questions relating to General Health, with 101
associated modalities of these questions; 24 questions relating to Depression, with 
154 associated modalities of these questions; and 7 questions relating to Demographics,
with 44 associated modalities of these questions.  

\begin{figure*}[h]
\centering
\includegraphics[width=10cm]{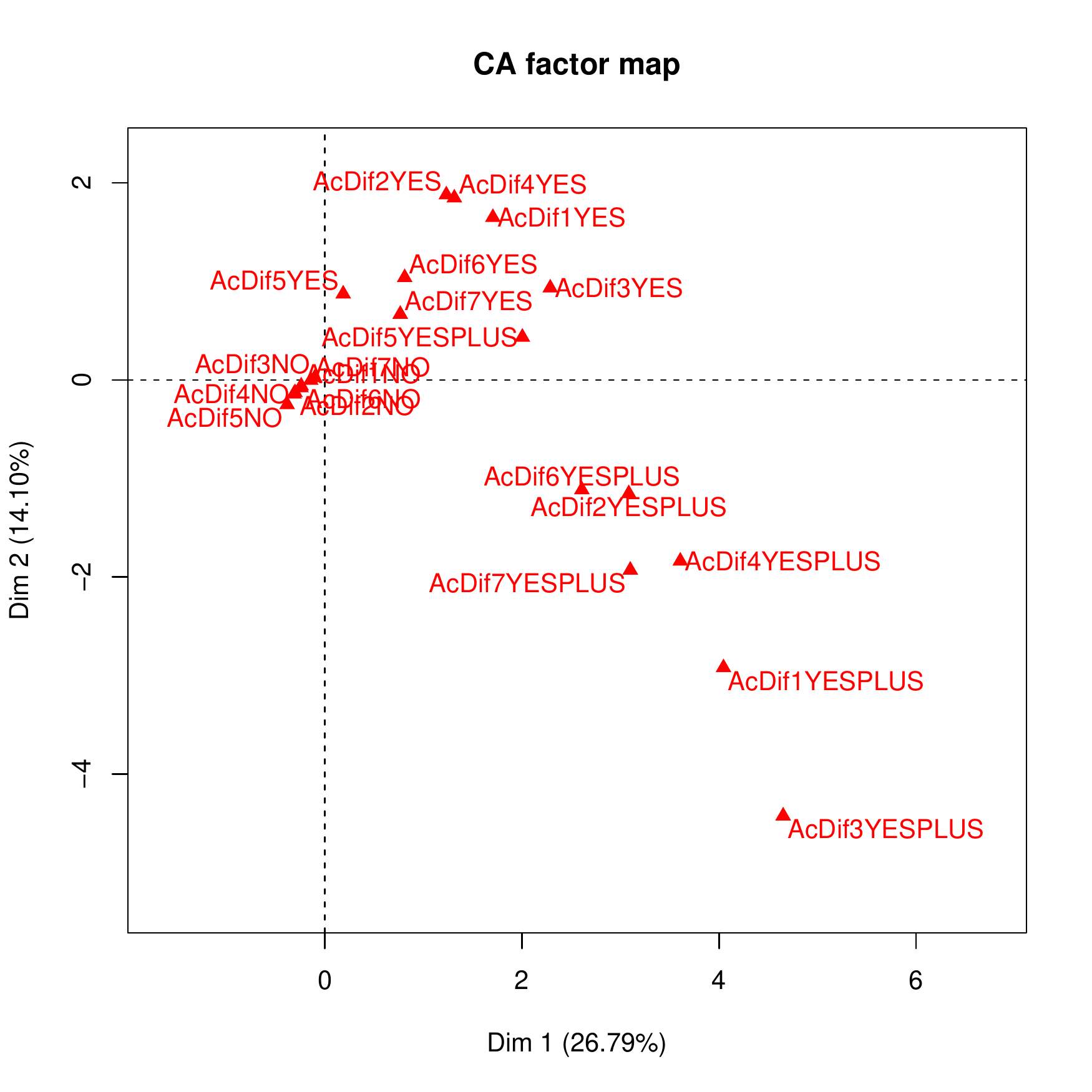}
\caption{This display of the modalities of the main, as opposed to supplementary,
question modalities is exactly the same factor space mapping as is displayed in 
Figures \ref{figcanew1} and \ref{figcanew2}.  The original, unmodified, inertias
are shown.  What is very clear is that the ``A lot of difficulty'' modalities are 
in the lower right quadrant, whereas the ``No difficulty at all'' modalities are 
close to the origin.}
\label{figcanew3}
\end{figure*}

\begin{figure*}[h]
\centering
\includegraphics[width=10cm]{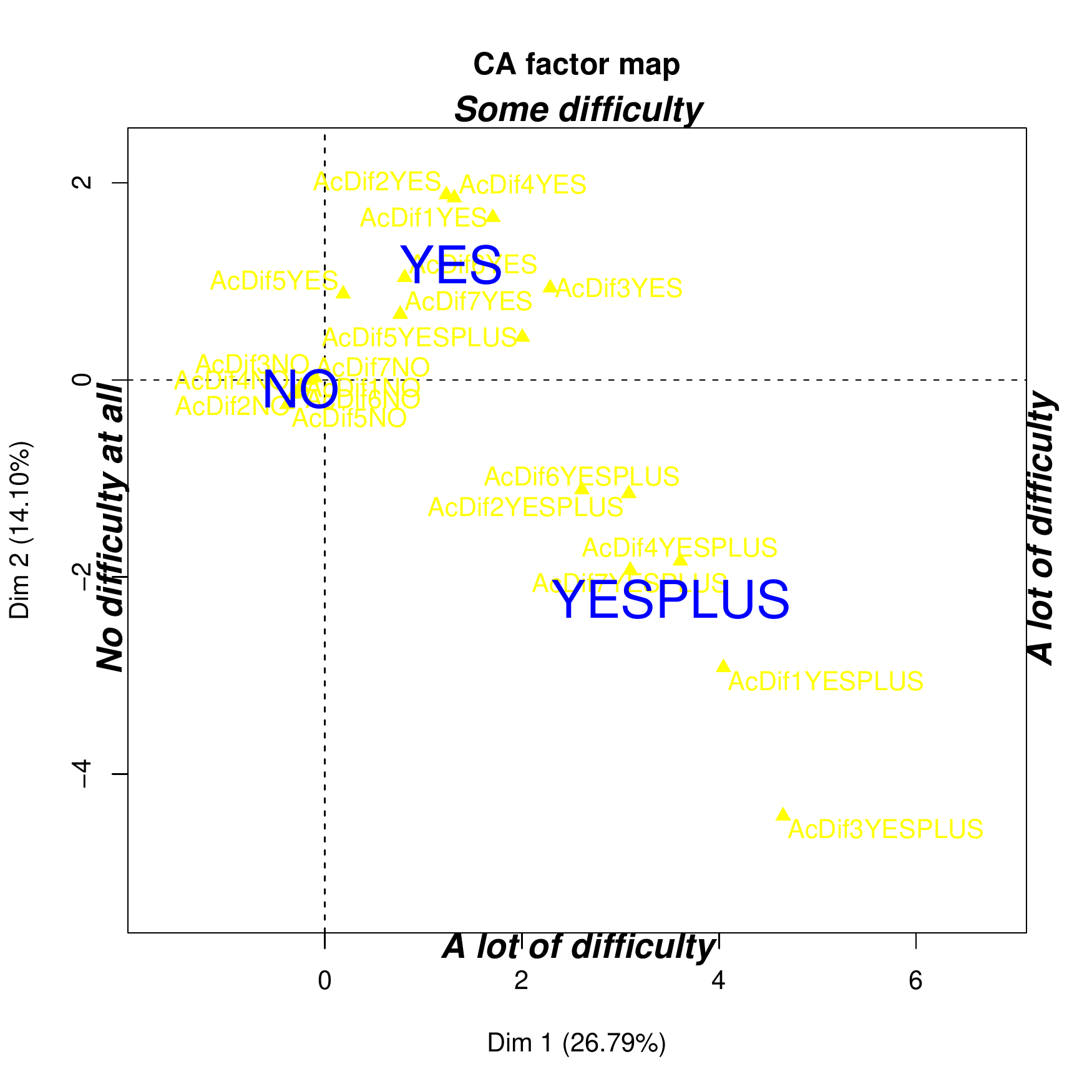}
\caption{With reference to Figure \ref{figcanew3}, here the centres of gravity, 
i.e.\ the means, of the three groups or clusters, are shown.  The modalities 
displayed in Figure \ref{figcanew3} are here shown in yellow.  Energetic and 
active is labelled with {\tt NO}; moderate difficulty in activity is labelled with 
{\tt YES}, and great difficulty indicated for activity is labelled with {\tt YESPLUS}.}
\label{figcanew4}
\end{figure*}

Figure \ref{figcanew3} is a display that is very helpful for our interpretation.  
It does use the original, unmodified, inertias, relative to the two previous 
displays of this principal factor plane.  As noted above, question AcDif5 relates 
to ``gardening, household repair''.   In Figure \ref{figcanew3}, the modality 
{\tt AcDif5YESPLUS}, i.e.\ the ``Yes, a lot of difficulty'' modality is close,
in location, to all the modalities of questions that indicated ``Yes, some difficulty''.
For the present, let us group together that modality, {\tt AcDif5YESPLUS}, with the
{\tt AcDif2YES}, i.e. ``Yes, some difficulty'', modalities.  Figure \ref{figcanew4} 
displays Figure \ref{figcanew3} a little differently.  It may be noted that in 
Figure \ref{figcanew3}, attention should be directed only at cosine squared values 
that are greater than 0.5.  That is to say, there is some non-negligible correlation 
with the axes. The practical cosine squared, hence correlation, with the first axis, 
consists of three ``NO'' and two ``YESPLUS'' response modalities.  For the second
axis, the highest cosine squared value is 0.4.  The contributions also need to 
be seen, in relation to Figure \ref{figcanew3}.  For axis 1, the highest contributions, 
greater than 8.6\%, are all ``YESPLUS'' modalities.   For axis 2, the highest valued
contributions are for three of the ``YES'' modalities that are greater than 10.9\%.  
Figure \ref{figcanew4} further illustrates these findings. 

\begin{figure*}[h]
\centering
\includegraphics[width=10cm]{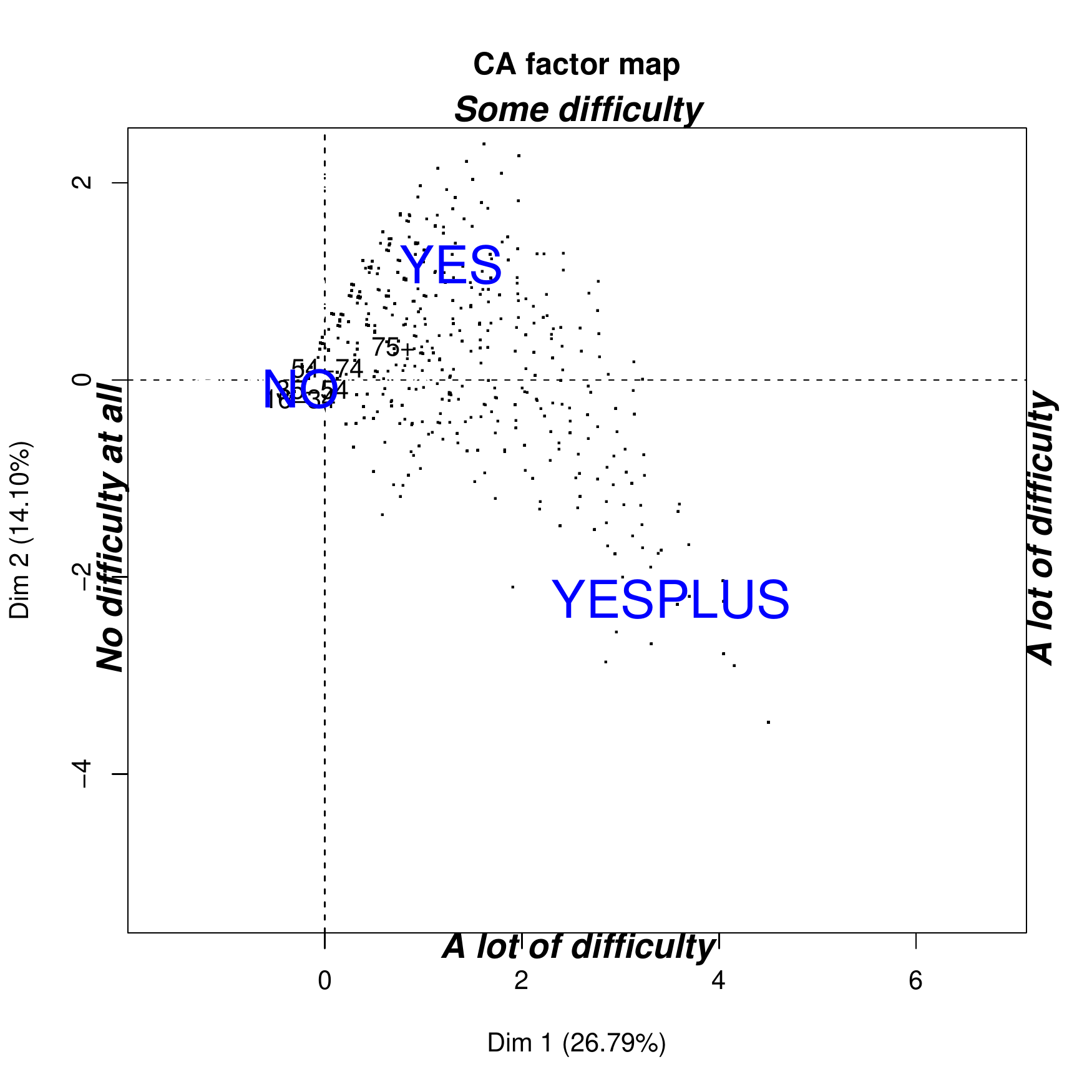}
\caption{Similarly to Figures \ref{figcanew3} and \ref{figcanew4}, here the 
respondents are shown with a dot for clarity of display rather than a label.
The means of the three group that we are interested in here, are shown.  Also 
the mean ages of respondents are plotted.}
\label{figcanew5}
\end{figure*}

Next, in Figure \ref{figcanew5}, just as in previous figures, but more explicitly 
shown here is the display of the three groupings, label {\tt NO}, ``No, No difficulty 
at all'' in relation to movements and activities, and the means of ages 16--34, 34--54, 
and 54--74 are all close to the origin, and the {\tt NO}, ``No, no difficulty at all''
response.  The very aged respondents tend towards the {\tt YES}, ``Yes, some difficulty''
response.  

%

We recall that we are dealing here with these variables: 19 General Health questions, 
comprising modalities 1 to 101; 7 Activities questions, comprising modalities 
102 to 126; 24 Depression questions, comprising modalities 127 to 280; and 7
Demographics questions, comprising modalities 281 to 324.  Now, of these, the 
supplementary modalities include four modalities from the first two question sets 
(i.e., General Health, Activities), taken as supplementary modalities because of 
their sparsity -- with total numbers of responses from the 7403 respondents being 
1, 2, 16, 2. Other supplementary modalities are all modalities from the General
Health, Depression and Demographics questions, comprising 19, 24, and 7 questions,
and these then comprise, respectively, 101, 154 and 44 modalities.  Adding the 
latter gives 299 modalities, and then with 4 (the sparse ones) with them, this
leads to 303 modalities here being supplementary modalities.  The main variables
here are the Activities, with 7 questions, comprising 25 modalities.  In the 
following we direct attention to the supplementary modalities. 

It was found that only the General Health questions were relevant for the top 
ranked cosines of all these modalities, in all 303, with the first factor.  The 
positive (tending towards: ``A lot of difficulty'') first factor projections were
the case for the 1st, 2nd, 5th, 7th and 10th of the top ranked cosines. 
The negative (tending towards: ``No difficulty at all'') first factor projections 
were the case for the 3rd, 4th, 6th, 8th and 9th of the top ranked cosines.  

The question for the 1st and 4th response modalities in the list of top ranked 
cosines with the first axis: 
 ``And how much does your health now limit you with these activities... [activities 
you might be doing during a typical day]''.

The question with modality responses that was the case for the 2nd, 5th and 8th 
of the top ranked cosines: 
``And how much does your health now limit you with this activity
  [in climbing several flights of stairs]''.  

The question for the 6th and 7th of the top ranked cosines is related to 
climbing several flights of stairs.
 
For the 3rd of the top ranked cosines, there was this question: 
``Does your health now limit you in moderate activities such as moving a
  table, pushing a vacuum cleaner, bowling or playing golf''.

Finally, for the 9th and 10th of the ten top ranked cosines with the first factor, 
there was this question: 
``And during the past 4 weeks, were you limited in the kind of work or other
  activities as a result of your physical health''.

In summary, we can conclude the following here.  Firstly, we have a planar, 
2-dimensional mapping that explains essentially all of the information content 
in the data.  Secondly, having focused attention here on action and activities, 
it is found that the first axis, hence the alignment associated with it, is the 
predominant interpretative outcome.   As noted, this focusing on activities was
both to be open to the increasing flow of secondary, contextual data that we 
can expect, arising out of modern technology.  Such technology is often labelled
Internet of Things or smarter cities.  It is wanted also, to consider a 
``No, no difficulty at all'' outlook in relation to activities and physical action, 
movement and motion. That has been seen in Figure \ref{figcanew1}.  Directly from
this there can follow the clustering or characterizing of the respondents. Finally, 
the most influential and relevant questions in the surveying that underlies the
data source were listed, and it was then noted which these were, what question 
set they were related to, and their ranking.  

%

What we also consider an important outcome is this: baselining, i.e.\ contextualizing,
against the ``No, No difficulty at all'' activity modality, cf.\ section 
\ref{sect2}.   In the ``No, No difficulty at all'' perspective, we have a clear
and coherent reference group for what we can label as mental capital.

\section{The Challenges and Opportunity in Big Data Analytics: Analytics
Focus and Contextualization}

Let us consider practical aspects of accounting for contextual relationships.
The context could provide the baseline or benchmark for what the data represents.
Perhaps one could present a case for the domain from which the data comes, 
being contrasted with some other domain.  That latter can be quite formal, 
and algorithmically well specified, for example, let us consider, in the 
long established methodology of canonical correlation analysis.  Below, however,
we will just look at meaningful structural relationships between more than one 
focus of interest, and relate these, quite briefly, to ``field'' as a focus 
of interest, and ``homology'' as an association linkage between certain properties
in these fields.  Those terms, field and homology, come from the work of eminent
social scientist, Pierre Bourdieu.  

In \cite{lebaron}, p.\ 43, there is this citation of Bourdieu:
``I make extensive use of correspondence analysis, in preference to multivariate 
regression, for instance, it is because correspondence analysis is a relational 
technique of data analysis whose philosophy corresponds exactly to what, in my
view, the reality of the social world is. It is a technique which `thinks' in 
terms of relation, as I try to do precisely in terms of field.'' 

\subsection{Towards Behaviour and Activity Analytics, for Mental Health}
\label{sect21} 

It is noted in \cite{kleinman} 
how relevant and important mental health 
is, given the integral association with physical health.  From \cite{kleinman} 
there is the following: ``... parity between mental and physical health 
conditions remains a distant ideal''.  ``The global economy loses about \$1 trillion
every year in productivity due to depression and anxiety''.  ``Next steps include 
... integration of mental health into other health and development sectors''.  

In \cite{cooper}, 
page 4, under the heading of ``Five Ways to Wellbeing'', 
reference is made to ``mental capital and wellbeing''.  On page 14, a section is 
entitled ``The `mental capital' values of the outdoors''.  


In \cite{murtagh2010}, 
and further described in \cite{murtagh2017}, 
an ontology is
assumed or created for the domain of application.  This leads to the main variables
in the analysis being taken from a high level in the taxonomic hierarchy, and then 
using lower level variables (in the taxonomic hierarchy) as supplementary variables. 
Such variables can be discipline-related, so that there is quite a natural 
taxonomic hierarchy at issue.   
This is at issue and to be further pursued in this work, relating to content-based
and qualitative analytics of published research literature: 
\cite{murtaghorlovmirkin}.  
Because higher order concepts are being employed for the primary and main attributes,
relative to lower order concepts that enter as supplementary attributes, it directly 
follows from the smaller number of higher order concept attributes that the inherent
factor space dimensionality is reduced.  In this sense, therefore, following from
the metaphor of a microscope, there is the ability to helpfully focus in on the 
interpretation of the data.

\subsection{Employing Big Data to Contextualize Sampled Data, with a Proposal 
for Lifestyle Analytics}

In the main analyses carried out in this article, it was noted that a motivation 
was to assume quantifiable action- and activity-related data to have a central and 
baseline role in the analytics.  Arising out of this perspective, let us consider 
contexts where data is voluminous and varied. 

Consider the essential requirements for contextualizing Big Data analytics.  
A major challenge in Big Data analytics is the bias due to self selection.  The 
consequences and repercussions for statistical sampling in Big Data analytics are 
discussed in \cite{keiding}.  
In the contribution to the discussion 
in \cite{keiding}, 
we point to field and homology analytics, following 
the work of eminent social scientist Pierre Bourdieu. We also point to the 
calibrating potential of Big Data.  In general, and in the work presented
here in this article, in practical settings, the aim is to determine the most revealing 
coupling of mainstream data and context.  This is technically processed in 
Correspondence Analysis through use of the main and the supplementary data 
elements, i.e., individuals or objects, attributes and modalities.  

We propose that analytics based on Bourdieu's work, based on MCA, (encompassing 
e.g.\ field, homology, habitus, etc.) 
should be a main analytics approach in many current areas of work, including smarter 
cities, analytics of Internet of Things, security and forensics (including trust and 
identity), Big Data, etc.   

\subsection{Lifestyle Analytics Requires Contextualization}

In \cite{mcclean} 
relating to activities of daily living (using the acronym, ADLs) 
of the elderly and those in poor health, there is use made of ``contextual 
information from uncertain sensor data''.  There is this: ``... our algorithm 
learns directly from incomplete data, and inhabitants' behavioural patterns are 
characterized using the learned probability distribution over various activities.
The model is used to infer the activities, and the inhabitants who have carried 
them out.''  For example, the monitoring of a kettle boiling water and perhaps 
some other sensor, these sensor data are used to model and predict some aspects
of ``making a tea with milk and sugar'', ``making coffee with milk'', etc. 
Further predictive models are at issue in \cite{mcclean2}, 
leading to a mobile
app that will trigger personalized reminders.  

The foregoing examples are such that sensor systems provide contextual data, 
as also do movement and other types of activities, with the main attention given 
to health and mental well-being.  

\subsection{Other Vantage Points: Social Activity in the Context of Health, and 
Health in the Context of Social Activity, and Analogously for Social Media.}

An objective of this article has been to contextualize the analysis of large and 
multi-faceted data sources.  For example, there may be
health research in the context of social characteristics.  Also there may 
be social research in the context of health characteristics.  
In \cite{murtagh2016b}, 
and as reported above in section \ref{sect4}, 
there is use of an 
adult psychiatric morbidity survey, 2007 data in England, encompassing a 
household survey, neurotic symptoms, common mental disorders, and 
socio-demographics.  Another large application to Twitter sources, with 
approximately 55 million tweets per annum, is at issue in the work of 
\cite{murtagh2016a}.  
This can be social media, i.e.\ Twitter, research in 
the context of social and 
individual characteristics, and it can be social and individual activity 
research in the context of social media.  The context of social media implies
being well expressed and well represented by this form of social media.

\section{Conclusions}

In this work, we have sought to provide perspectives and focus for the analytics
of behaviours and activities, that can be related to the determining of forms of 
mental capital.  We discovered in our analyses how a few forms of activity play 
a very central role in such contexts as health and depression, and demographic 
characteristics.  

This work was based on one year's surveying, and it is planned to be further 
pursued by having a pre-prepared approach to investigate trends over time and 
other patterns that may be discovered in the data.  

In line with our discussion in \cite{keiding}, 
additionally we will be responding 
to the problem of bias in big data sampling and related representativity. 
We also seek to bridge the data with decision-making information.  In the sense of 
the latter, we are bridging data analytics with position-taking, i.e.\ decision-making.

One final point is to note that factor space mapping methods can render interpretation 
difficult, when the number of variables or of modalities of the variables are very 
large.  One reason for this is just how mapping and display becomes when there is a 
lot to be analysed, interpreted and understood simultaneously.  However it may be the 
case that Multiple Correspondence Analysis or any related methods can be taken as a 
preliminary or intermediate technique in order to produce, or set up the data for, 
other methods. Examples of the latter often include clustering, and statistical 
modelling or other methods may be the case.  That simply leads to the conclusion, 
that Multiple Correspondence Analysis or any related methods can be very appropriate
as an early stage in the analytics carried out.  Such methods may be a means of 
having valuable, relevant and new perspectives and vantage points on the data.  

\bibliographystyle{jimis-en}
\bibliography{ContextualizingDataAnalyticsRefs}

\end{document}